\documentclass[9pt,conference]{IEEEtran}

\usepackage{amsfonts}
\usepackage{bm}
\usepackage{graphicx}
\usepackage{lipsum}
\usepackage{xcolor}

\usepackage{cite}

\ifCLASSINFOpdf
\else
\fi
\usepackage[cmex10]{amsmath}
\usepackage[tight,footnotesize]{subfigure}
\usepackage{url}
\begin{document}
%
% paper title
% can use linebreaks \\ within to get better formatting as desired
\title{Spatio-temporal spike and slab priors for MMV problems}

% author names and affiliations
% use a multiple column layout for up to three different
% affiliations
\author{\IEEEauthorblockN{Michael Riis Andersen, Ole Winther \& Lars Kai Hansen}
\IEEEauthorblockA{DTU Compute, Technical University of Denmark\\
DK-2800 Kgs. Lyngby, Denmark\\
Email: \{miri, olwi, lkh\}@dtu.dk}}

% conference papers do not typically use \thanks and this command
% is locked out in conference mode. If really needed, such as for
% the acknowledgment of grants, issue a \IEEEoverridecommandlockouts
% after \documentclass

% for over three affiliations, or if they all won't fit within the width
% of the page, use this alternative format:
% 
%\author{\IEEEauthorblockN{Michael Shell\IEEEauthorrefmark{1},
%Homer Simpson\IEEEauthorrefmark{2},
%James Kirk\IEEEauthorrefmark{3}, 
%Montgomery Scott\IEEEauthorrefmark{3} and
%Eldon Tyrell\IEEEauthorrefmark{4}}
%\IEEEauthorblockA{\IEEEauthorrefmark{1}School of Electrical and Computer Engineering\\
%Georgia Institute of Technology,
%Atlanta, Georgia 30332--0250\\ Email: see http://www.michaelshell.org/contact.html}
%\IEEEauthorblockA{\IEEEauthorrefmark{2}Twentieth Century Fox, Springfield, USA\\
%Email: homer@thesimpsons.com}
%\IEEEauthorblockA{\IEEEauthorrefmark{3}Starfleet Academy, San Francisco, California 96678-2391\\
%Telephone: (800) 555--1212, Fax: (888) 555--1212}
%\IEEEauthorblockA{\IEEEauthorrefmark{4}Tyrell Inc., 123 Replicant Street, Los Angeles, California 90210--4321}}

% use for special paper notices
%\IEEEspecialpapernotice{(Invited Paper)}

% make the title area
\maketitle

%\begin{abstract}
%\boldmath
%The abstract goes here.
%\end{abstract}
% IEEEtran.cls defaults to using nonbold math in the Abstract.
% This preserves the distinction between vectors and scalars. However,
% if the conference you are submitting to favors bold math in the abstract,
% then you can use LaTeX's standard command \boldmath at the very start
% of the abstract to achieve this. Many IEEE journals/conferences frown on
% math in the abstract anyway.

% no keywords

% For peer review papers, you can put extra information on the cover
% page as needed:
% \ifCLASSOPTIONpeerreview
% \begin{center} \bfseries EDICS Category: 3-BBND \end{center}
% \fi
%
% For peerreview papers, this IEEEtran command inserts a page break and
% creates the second title. It will be ignored for other modes.
\IEEEpeerreviewmaketitle

% MRA definitions
\newcommand{\A}{\mathbf{A}}
\newcommand{\E}{\mathbf{E}}
\newcommand{\I}{\mathbf{I}}
\newcommand{\X}{\mathbf{X}}
\newcommand{\Y}{\mathbf{Y}}

\newcommand{\e}{\mathbf{e}}
\newcommand{\m}{\mathbf{m}}
\newcommand{\x}{\mathbf{x}}
\newcommand{\y}{\mathbf{y}}
\newcommand{\z}{\mathbf{z}}
\newcommand{\Z}{\mathbf{Z}}

\newcommand{\N}{\mathcal{N}}

\begin{abstract}
%\boldmath
We are interested in solving the multiple measurement vector (MMV) problem for instances, where the underlying sparsity pattern exhibit spatio-temporal structure motivated by the electroencephalogram (EEG) source localization problem. We propose a probabilistic model that takes this structure into account by generalizing the structured spike and slab prior and the associated Expectation Propagation inference scheme. Based on numerical experiments, we demonstrate the viability of the model and the approximate inference scheme.
\end{abstract}

\section{Introduction}
% no \IEEEPARstart
The multiple measurement vector problem (MMV) \cite{Cotter05sparsesolutions} is given by:
\begin{align} \label{eq:linearinverseproblem}
\Y = \A\X + \E,
\end{align}
where $\A \in \mathbb{R}^{N \times D}$ is the forward matrix, $\Y \in \mathbb{R}^{N\times T}$ is the measurement matrix, $\X = \begin{bmatrix}\x_1&\x_2&\hdots&\x_ T\end{bmatrix} \in \mathbb{R}^{D\times T}$ is the desired solution and $\E \in \mathbb{R}^{N\times T}$ is a matrix of corruptive noise. We are interested in finding sparse solutions to eq. \eqref{eq:linearinverseproblem} in the ill-posed regime, where $N < D$. Furthermore, the sparsity pattern of $\X$ is assumed to have certain structural properties. In particular, we are considering problems where the sparsity pattern exhibit spatio-temporal structure as in EEG source localization \cite{baillet2001a,a2012a} or in background subtraction in computer vision \cite{cevher2009a}. Let $\z_t$ be an indicator for the support of $\x_t$, i.e. $\z_{t} = \mathbb{I}\left[\x_{t} \neq 0\right]$, then $\z_t$ is assumed to be spatially correlated. Furthermore, we assume that the support vectors $\z_1, \z_2, ..., \z_T$ slowly evolve through time as well - rendering the joint sparsity assumption invalid \cite{berg2010a}.
\\
\\
The main contribution of this work is to propose a model for spatio-temporal sparsity patterns by extending the structured spike and slab prior \cite{NIPS2014_5464} to account for temporal evolution of the sparsity pattern as well. Furthermore, we demonstrate the benefits of the model through numerical experiments. 

\subsection{Related work}
The field of structured sparsity has received a great deal of attention in the recent years. In this section we highlight some of the related work, but this list is by no means exhaustive. The LASSO-community have introduced the Group and Graph LASSO methods, which generalize the standard $\ell_1$-norm minimization approach to promote different kinds of structured sparsity \cite{jacob2009a}. In the probabilistic setting, the standard workhorse for sparsity is the so-called spike and slab prior \cite{mitchell1988a}. This has also been generalized to model group sparsity \cite{hernandez2013a} and cluster sparsity \cite{yu2011a}. In the context of compressed sensing \cite{donoho2006a}, Cevher et al. \cite{cevher2009a} used a Markov random field to enforce spatially correlated sparsity patterns, whereas Ziniel et al. used binary Markov chains to model temporally correlated sparsity patterns \cite{ziniel2010a}.

\section{The structured spike and slab prior}
In this section we briefly introduce the conventional spike and slab prior \cite{mitchell1988a} and the structured spike and slab prior \cite{NIPS2014_5464} before we move on to the spatio-temporal spike and slab prior on  the next section. The conventional spike and slab prior decomposes each $x_{i,t}$ as a product of a binary variable $z_{i,t}$ and a real number $c_{i,t}$, i.e. $x_{i,t} = z_{i,t} c_{i,t}$, where $z_{i,t} \sim \text{Ber}\left(p_0\right)$ and $c_{i,t} \sim \N\left(0, \tau_0\right)$ for $i \in \left\lbrace 1, 2, .., D\right\rbrace$ and $t \in \left\lbrace 1, 2, .., T\right\rbrace$. The structured spike and slab prior generalized this formulation by imposing structure on the binary variable for each time $t$ as follows
\begin{align} \label{eq:prior_z}
p(\z_t\big|\phi\left(\bm{\gamma}_t\right)) &= \prod_{i=1}^D \text{Ber}\left(z_{i,t}\big|\phi\left({\gamma}_{i,t}\right)\right),\\
p(\bm{\gamma}_t) &= \N\left(\bm{\gamma}_t\big|\bm{\mu}_t, \bm{\Sigma}_t\right),
\end{align}
where the Bernoulli probabilities are parametrized using the standard normal CDF $\phi: \mathbb{R} \rightarrow \left(0,1\right)$. The hyperparameters $\bm{\mu}_t$ and $\bm{\Sigma}_t$ encode the prior belief of the support for time $t$.  Specifically, the prior mean value $\bm{\mu}_t$ controls the prior belief of the number of non-zero variables and the covariance matrix $\bm{\Sigma}_t$ determines the prior correlation of the support at time $t$. Thus, we can impose structure on the binary support variables $\z_t$ by means of imposing generic covariance functions on $\bm{\gamma}$. For example, say we choose $\Sigma_{i,j}$ to be the squared exponential covariance function, then the resulting prior distribution will promote sparsity patterns where neighbouring support variables have the same state. Under the other hand, when $\bm{\Sigma}$ is diagonal, we recover the independent spike and slab prior.
\\
\\
The marginal prior probability of the $x_{i,t}$ being non-zero is given by
\begin{align}
p(z_{i,t} = 1) &= \int p(z_{i,t} = 1\big|\gamma_{i,t})p(\gamma_{i,t}) \text{d} \gamma_{i,t}\nonumber\\
&= \int \phi(\gamma_{i,t}) \N\left(\gamma_{i,t}\big|\mu_{i,t}, \Sigma_{ii,t}\right) \text{d} \gamma_{i,t}\nonumber\\
 &= \phi\left(\frac{\mu_{i,t}}{\sqrt{1+\Sigma_{ii,t}}}\right).
\end{align}
Thus, if the prior on $\bm{\gamma}_t$ has zero mean, then the prior belief of $p(z_{i,t})$ is unbiased, i.e. $p(z_{i,t}) = 0.5$. On the other hand, if $\mu_{i,t}$ is negative, the prior belief of $z_{i,t}$ is biased towards zero and vice versa.

\section{The spatio-temporal spike and slab prior}
In this section we describe the temporal extension of the structured spike and slab prior. Instead of considering $\bm{\mu}_t$ and $\bm{\Sigma}_t$ as fixed hyperparameters, we propose to impose a prior on $\bm{\Gamma} = \begin{bmatrix}
\bm{\gamma}_1&\bm{\gamma}_2&\hdots&\bm{\gamma}_T \end{bmatrix}$ to model problems where the support of the solution $\X$ changes over time. In particular, we impose a first order process Markov process on $\bm{\Gamma}$ to model the slowly changing sparsity pattern
\begin{align} \label{eq:prior_gamma}
p\left(\bm{\gamma}_t\big|\bm{\gamma}_{t-1}\right) = \N\left(\bm{\gamma}_t\big|\left(1-\alpha\right)\bm{\mu}_0 + \alpha \bm{\gamma}_{t-1}, \beta \bm{\Sigma}_0\right),
\end{align}
where the hyperparameters $\alpha$ and $\beta$ control the temporal correlation and the ''innovation'' of the process, respectively. Furthermore, we assume that the prior distribution on $\bm{\gamma}_1$ is given by
\begin{align}
p(\bm{\gamma}_1) = \N\left(\bm{\gamma}_1\big|\bm{\mu}_0, \bm{\Sigma}_0\right).
\end{align}
Under these assumptions the marginal distribution of $\bm{\gamma}_2$ becomes
\begin{align}
p(\bm{\gamma}_2) &= \int p\left(\bm{\gamma}_2\big|\bm{\gamma}_{1}\right)p(\bm{\gamma}_1) \text{d} \bm{\gamma}_1\nonumber\\
&= \N\left(\bm{\gamma}_2\big|\bm{\mu}_0, \left(\alpha^2 + \beta\right)\bm{\Sigma}_0\right).
\end{align}
Therefore, it follows by induction that if $\alpha$ and $\beta$ satisfy $\alpha^2 + \beta = 1$, then the marginal distribution of $\bm{\gamma}_t$ is $p(\bm{\gamma}_t) = \N\left(\bm{\mu}_0, \bm{\Sigma}_0\right)$ for all $t$. Furthermore, we also see that for $\alpha = 1$ and $\beta = 0$, the prior reduces to the structured spike and slab prior in the joint sparsity setting. In the other extreme, at $\alpha = 0$ and $\beta = 1$, the prior reduces to the structured spike and slab prior in the time-independent setting.  Hence, the spatio-temporal spike and slab prior can be seen as a generalization of the two extreme cases.
\\
\\
This choice of model is also motivated by the fact that the first order structure in the temporal dimension gives rise to an inference scheme that scales linearly in the number of time steps $T$ as we will see in the next section.

\section{Bayesian Inference using the spatio-temporal spike and slab prior}
The goal of this section is to describe an inference procedure for solving the problem in eq. \eqref{eq:linearinverseproblem} using the proposed prior in a fully Bayesian setting. We combine the spatio-temporal spike and slab prior with a time-independent isotropic Gaussian noise model of the form
\begin{align}
p(\Y\big|\X) = \prod_{t=1}^T \N\left(\y_t\big|\A\x_t, \sigma^2_0 \I\right).
\end{align} 
This gives rise to the following joint distribution
\begin{align}
\!p(\Y, \X, \Z, \bm{\Gamma})\! =\! &\underbrace{\prod_{t=1}^T \N\left(\y_t\big|\A\x_t, \sigma^2_0 \I\right)}_{f_1\left(\X\right)}\nonumber\\
&\underbrace{\prod_{t=1}^T \prod_{i=1}^D \left[(1-z_{i,t})\delta(x_{i,t}) + z_{i,t}\N\left(x_{i,t}\big|0, \tau_0\right)\right]}_{f_2\left(\X, \Z\right)}\nonumber\\
&\underbrace{\prod_{t=1}^T\prod_{i=1}^D \text{Ber}\left(z_{i,t}\big|\phi\left(\gamma_{i,t}\right)\right)}_{f_3\left(\Z, \bm{\Gamma}\right)}\nonumber\\
&\underbrace{\N\!\left(\bm{\gamma}_1\big|\bm{\mu}_0, \bm{\Sigma}_0\right)\! \prod_{t=2}^T \N\!\left(\bm{\gamma}_t\big|\!\left(1\!-\!\alpha\right)\!\bm{\mu}_0 + \alpha \bm{\gamma}_{t-1}, \beta \bm{\Sigma}_0\right)}_{f_4\left(\bm{\Gamma}\right)} \label{eq:full_joint_model}
\end{align}
The desired posterior distribution $p(\X, \Z, \bm{\Gamma}\big|\Y)$ is obtained from Bayes' Rule \cite{bishop2006a}. Unfortunately, this posterior distribution is intractable due to the product of mixtures and hence, we have to settle for approximate inference. Specifically, we use Expectation Propagation \cite{minka2013a, opper2000gaussian, seeger2009a} for approximate inference by extending the proposed inference scheme in \cite{NIPS2014_5464}. 

\subsection{Approximate Inference using Expectation Propagation}
Expectation propagation (EP) is an iterative deterministic method for approximating probability distributions using simpler distributions from the exponential family. As indicated in eq. \eqref{eq:full_joint_model}, the exact posterior can be decomposed as follows
\begin{align}
p(\X, \Z, \bm{\Gamma}\big|\Y) =  &\frac{1}{Z}\prod_{t=1}^T f_{1,t}\left(\x_t\right)\prod_{t=1}^T\prod_{i=1}^D f_{2, i, t}\left(x_{i,t}, z_{i,t}\right)\nonumber\\
&\prod_{t=1}^T\prod_{i=1}^D f_{3,i,t}\left(z_{i,t}, \gamma_{i,t}\right) \prod_{t=1}^T f_{4,t}\left(\bm{\gamma}_t\right), \label{eq:full_joint_model_factorized}
\end{align}
where $Z = p\left(\Y\right)$ is the normalization constant. Moreover, note that each factor in the decomposition only depends on a subset of the variables in the model, i.e. $f_{2,i,t}$ depends only on the variables $x_{i,t}$ and $z_{i,t}$ and so on and so forth. The EP framework takes advantage of this decomposition by approximating each factor in eq. \eqref{eq:full_joint_model_factorized} with a distribution from the exponential family. First we describe the functional form of the approximation and then we briefly explain how to estimate the parameters of the approximation using the EP algorithm.
\\
\\
Let $\tilde{f}_{1,t}$ denote the approximation of $f_{1,t}$ etc. First, we note that each of the factors in the first term, i.e. $f_{1,t}$ for all $t$, are already a member of the exponential family and hence does not have to be approximated. Therefore, for each $t$ we have
\begin{align}
\tilde{f}_{1,t}\left(\x_t\right) = \N\left(\x_t\big|\tilde{\m}_{1,t}, \tilde{\bm{V}}_{1,t}\right),
\end{align}
where the parameters are determined by $\tilde{\bm{V}}^{-1}_{1,t}\tilde{\bm{m}}_{1,t} = \frac{1}{\sigma^2_0}\A^T\y_t$ and $\tilde{\bm{V}}^{-1}_{1,t} = \frac{1}{\sigma^2_0}\A^T \A$. Note that the exact term $f_{1,t}$ is a distribution on $\y_t$ conditioned on $\x_t$, whereas the approximate term $\tilde{f}_{1,t}$ is a function of $\x_t$ that depends on $\y_t$ through $\tilde{\m}_{1,t}$ and $\tilde{\bm{V}}_{1,t}$ etc. Next, we turn to the factors in the second term, i.e. $f_{2, i,t}$. Since each of these factors depends on $x_{i,t}$ and $z_{i,t}$, we choose $\tilde{f}_{2,i,t}$ to be
\begin{align}
\tilde{f}_{2,i,t} = \N\left(x_{i,t}\big|\tilde{m}_{2,i,t}, \tilde{V}_{2,i,t}\right)\text{Ber}\left(z_{i,t}\big|\phi\left(\tilde{\gamma}_{2,i,t}\right)\right),
\end{align} 
where $\tilde{m}_{2,i,t}$, $\tilde{V}_{2,i,t}$ and $\tilde{\gamma}_{2,i,t}$ have to determined using the EP algorithm. Based on similar arguments $\tilde{f}_{3,i,t}$ and $\tilde{f}_{4,t}$ are chosen as follows
\begin{align}
\tilde{f}_{3,i,t} &= \text{Ber}\left(z_{i,t}\big|\phi\left(\tilde{\gamma}_{3,i,t}\right)\right)\N\left(\gamma_{3,i,t}\big|\tilde{\mu}_{3, i, t}, \tilde{\Sigma}_{3,i,t}\right),\\
\tilde{f}_{4,t} &= \N\left(\bm{\gamma}_t\big|\tilde{\bm{\mu}}_{4,t}, \tilde{\bm{\Sigma}}_{4,t}\right).
\end{align}
Note that $f_{4,1}$ does not have to approximated either, it is simply $\tilde{f}_{4,1} = \N\left(\bm{\gamma}_1\big|\bm{\mu}_0, \bm{\Sigma}_0\right)$. Furthermore, note that the approximations to the factors $f_{4,t}$ for all $t$ do not factorize w.r.t. $\gamma_{t,1}, \gamma_{t,2}, ... $ in order to capture potentially strong correlations in the support. 
\\
\\
After specifying all the individual approximation terms, we derive the joint approximation of the desired posterior $p(\X, \Z, \bm{\Gamma}\big|\Y)$. Since the exponential family is closed under products, the approximate joint distribution has the following form
\begin{align}
Q\left(\X, \Z, \bm{\Gamma}\right) = &\prod_{t=1}^T \N\left(\x_t\big|\tilde{\m}_t, \bm{\tilde{V}}_t\right) \prod_{t=1}^T \prod_{i=1}^D \text{Ber}\left(z_{i,t}\big|\phi\left(\tilde{\gamma}_{i,t}\right)\right)\nonumber\\
&\prod_{t=1}^T \N\left(\bm{\gamma}_t\big|\tilde{\bm{\mu}}_t, \tilde{\bm{\Sigma}}_t\right) \label{eq:joint_approx}.
\end{align}
Let $\m_{2,t} = \left[\tilde{m}_{2,1,t}, \tilde{m}_{2,2,t}, \hdots, \tilde{m}_{2,D,t}\right]^T$ and $\bm{V}_{2,t} = \text{diag}\left(\tilde{V}_{2,1,t}, \tilde{V}_{2,2,t}, \hdots, \tilde{V}_{2,D,t}\right)$, and analogously for $\tilde{\bm{\mu}}_3$, $\tilde{\bm{\Sigma}}_3$ and $\bm{\gamma}_3$, then the parameters of the joint approximation are given by
\begin{align}
\tilde{\bm{V}}_{t} &= \left(\tilde{\bm{V}}^{-1}_{1,t} + \tilde{\bm{V}}^{-1}_{2,t}\right)^{-1}, \label{eq:joint_update_start}\\
\tilde{\m}_t &= \tilde{\bm{V}}_{t}\left(\tilde{\bm{V}}^{-1}_{1,t}\tilde{\m}_{1,t} + \tilde{\bm{V}}^{-1}_{2,t}\tilde{\m}_{2,t}\right),\\
\tilde{\bm{\Sigma}}_{t} &= \left(\tilde{\bm{\Sigma}}_{3,t} + \tilde{\bm{\Sigma}}_{4,t}\right)^{-1},\\\
\tilde{\bm{\mu}}_{t} &=  \tilde{\bm{\Sigma}}_{t}\left(\tilde{\bm{\Sigma}}^{-1}_{3,t}\tilde{\bm{\mu}}_{3,t} + \tilde{\bm{\Sigma}}^{-1}_{4,t}\tilde{\bm{\mu}}_{4,t}\right), \label{eq:joint_update_sigma}\\
\tilde{{\gamma}}_{i,t} &= \phi^{-1}\left[\left(\frac{\left(1-\phi(\tilde{\gamma}_{2,i,t})\right)\left(1-\phi(\tilde{\gamma}_{3,i,t})\right)}{\phi(\tilde{\gamma}_{2,i,t})\phi(\tilde{\gamma}_{3,i,t})} + 1\right)^{-1}\right].  \label{eq:joint_update_end}
\end{align}
The posterior covariance matrices $\tilde{\bm{V}}_t$ and $\tilde{\bm{\Sigma}}_t$ are (potentially) fully dense matrices, which makes the approximation able to cope with non-orthogonal forward matrices $\A$. 

\subsection{The Expectation Propagation Algorithm}
In this section we describe how to compute the parameters of the individual approximations using the EP algorithm. The EP algorithm works by updating each of the individual approximation terms one by one until convergence. Consider the update of the term $\tilde{f}_{a,i,t}$ for a given $a$, $i$ and $t$. The update is obtained by performing the following three steps of the EP algorithm. The first step is to remove the contribution of $\tilde{f}_{a,i,t}$ from the joint approximation in eq. \eqref{eq:joint_approx} by forming the so-called cavity distribution
\begin{align}
Q^{\backslash a,i,t} \propto \frac{Q}{\tilde{f}_{a,i,t}}.
\end{align}
In the next step we minimize the Kullbach-Leibler \cite{bishop2006a} divergence between $f_{a,i,t}Q^{\backslash a,i,t}$ and $Q^{a, t, \text{new}}$ w.r.t. $Q^{a, t,\text{new}}$. That is, we minimize $\text{KL}\left(\frac{1}{Z_{a,i,t}} f_{a,i,t}Q^{\backslash a,i,t}||Q^{a, t, \text{new}}\right)$, where $Z_{a,i,t}$ is the normalization constant of $f_{a,i,t}Q^{\backslash a,i,t}$. For distributions within the exponential family, minimizing this form of KL divergence amounts to matching moments between $f_{a,i,t}Q^{\backslash a,i,t}$ and $Q^{a,t,\text{new}}$ \cite{minka2013a}. Finally, the third and last step is to compute the new update of $\tilde{f}_{a,i,t}$ as follows 
\begin{align}
\tilde{f}_{a,i,t} \propto \frac{Q^{a,t,\text{new}}}{Q^{\backslash a,i,t}}.
\end{align}
After the individual approximation terms $\tilde{f}_{a,i,t}$ for all $i$ and $t$ for a given $a$ have been updated, the relavant parts of the joint approximation are updated using eq. \eqref{eq:joint_update_start}-\eqref{eq:joint_update_end}. To minimize the computational load, we use parallel updates of $\tilde{f}_{2,i,t}$ \cite{hernandez2013a} followed by parallel updates of $\tilde{f}_{3,i,t}$ rather than the conventional sequential update scheme. Furthermore, due to the fact that $\tilde{f}_2$ and $\tilde{f}_3$ factorizes w.r.t. both $i$ and $t$, we only need the marginals of the cavity distributions $Q^{\backslash a, i, t}$, which simplifies the computations. Computing the cavity distributions and matching the moments are straightforward. However, when matching the moments, we are required to evaluate of the zero'th, first and second order moment of the distributions of the form $\phi(\gamma_i)\mathcal{N}\left(\gamma_i\big|\mu_i, \Sigma_{ii}\right)$. Derivation of analytical expressions for these moments can be found in the appendix to chapter 3 in \cite{rasmussen2006a}.
\\
\\
The proposed EP algorithm is summarized in figure \ref{fig:algorithm}. The computational complexity of the algorithm is dominated by the matrix inversions in eq. \eqref{eq:joint_update_start} and \eqref{eq:joint_update_sigma}. However, when $N < D$, the covariance matrices $\tilde{\bm{V}}_{1,t}$ have low rank and hence, eq. \eqref{eq:joint_update_start} can be carried out in $\mathcal{O}\left(ND^2\right)$ using the Matrix Inversion Lemma \cite{petersen2012a}. Therefore, the resulting inference scheme scales as $\mathcal{O}\left(T D^3\right)$, i.e. it scales linearly in the number of measurement vectors $T$. 

\begin{figure}[t]
\fbox{\parbox[t][\height][t]{0.465\textwidth}{
\begin{itemize}
\item Initialize approximation terms $\tilde{f}_a$ for $a = 1, 2,3,4$ and $Q$
\item Repeat until stopping criteria
\begin{itemize}
\item For each $\tilde{f}_{2,i,t}$:
\begin{itemize}
\item Compute cavity distribution: $Q^{\backslash 2,i,t} \propto \frac{Q}{\tilde{f}_{2,i,t}}$
\item Minimize: KL$\left(f_{2,i,t}Q^{\backslash 2,i,t}\big|\big|Q^{2, t,\text{new}}\right)$ w.r.t. $Q^{\text{new}}$
\item Compute: $\tilde{f}_{2,i,t} \propto \frac{Q^{2,t,\text{new}}}{Q^{\backslash 2,i,t}}$ to update parameters $\tilde{m}_{2,i,t}, \tilde{v}_{2,i,t}$ and $\tilde{\gamma}_{2,i,t}$.
\end{itemize}
\item Update joint approximation parameters: $\tilde{\m}, \tilde{\bm{V}}$ and $\tilde{\bm{\gamma}}$ 
\item For each $\tilde{f}_{3,i,t}$:
\begin{itemize}
\item Compute cavity distribution: $Q^{\backslash 3,i,t} \propto \frac{Q}{\tilde{f}_{3,i,t}}$
\item Minimize: KL$\left(f_{3,i,t}Q^{\backslash 3,i,t}\big|\big|Q^{3,t,\text{new}}\right)$ w.r.t. $Q^{3,t,\text{new}}$
\item Compute: $\tilde{f}_{3,i,t} \propto \frac{Q^{3,t,\text{new}}}{Q^{\backslash 3,i,t}}$ to update parameters $\tilde{\mu}_{3,i,t}, \tilde{\sigma}_{3,i,t}$ and $\tilde{\gamma}_{3,i,t}$
\end{itemize}
\item Update joint approximation parameters: $\tilde{\bm{\mu}}, \tilde{\bm{\Sigma}}$ and $\tilde{\bm{\gamma}}$
\item For each $\tilde{f}_{4,t}$
\begin{itemize}
	\item Compute cavity distribution: $Q^{\backslash 4,t} \propto \frac{Q}{\tilde{f}_{4,t}}$
	\item Minimize: KL$\left(f_{4,t}Q^{\backslash 4,i}\big|\big|Q^{4, t,\text{new}}\right)$ w.r.t. $Q^{\text{new}}$
	\item Compute: $\tilde{f}_{4,t} \propto \frac{Q^{4,t,\text{new}}}{Q^{\backslash 4,t}}$ to update parameters $\tilde{m}_{4,t}, \tilde{v}_{4,t}$ and $\tilde{\gamma}_{4,t}$.
\end{itemize}
\item Update joint approximation parameters: $\tilde{\bm{\mu}}, \tilde{\bm{\Sigma}}$
\end{itemize}
\end{itemize}
} 
}
\caption{Proposed algorithm for approximating the joint posterior distribution over $\X, \Z$ and $\bm{\Gamma}$ conditioned on $\Y$.}
\label{fig:algorithm}
\end{figure}

\subsection{Tuning of hyperparameters}
The algorithm requires tuning of multiple hyperparameters for optimal performance. The Expectation Propagation framework provides a neat alternative to typical cross-validation schemes. Besides the approximation to the posterior distribution $P(\X, \Z, \bm{\Gamma}\big|\Y)$, EP also provides an approximation to the marginal likelihood $P(\Y)$, which is very useful for model selection and tuning of hyperparameters \cite{bishop2006a}. The exact marginal likelihood is obtained by marginalizing out $\X, \Z$ and $\bm{\Gamma}$ from the joint distribution in eq. \eqref{eq:full_joint_model}. The EP approximation to the marginal likelihood is obtained by substituting all the (scaled) individual approximation terms into the resulting formula. Finally, it is also possible to get closed form expression for the gradients of the marginal likelihood approximation w.r.t. to the hyperparameters \cite{seeger2009a, rasmussen2006a}, which allows efficient tuning of the hyperparameters. 
However, a detailed treatment of the marginal likelihood approximation and its gradient w.r.t. hyperparameters are out of scope for this extended abstract.

\section{Numerical Experiments}
In order to investigate the properties of the proposed algorithm, we have designed and conducted two numerical experiments. The first experiment addresses the reconstruction performance of the algorithm, whereas the second experiment investigate the algorithm's robustness towards coherent forward models.

\subsection{Experiment 1}
\begin{figure}[!t] 
\centering
%\subfigure[]{
\includegraphics[width=0.48\textwidth]{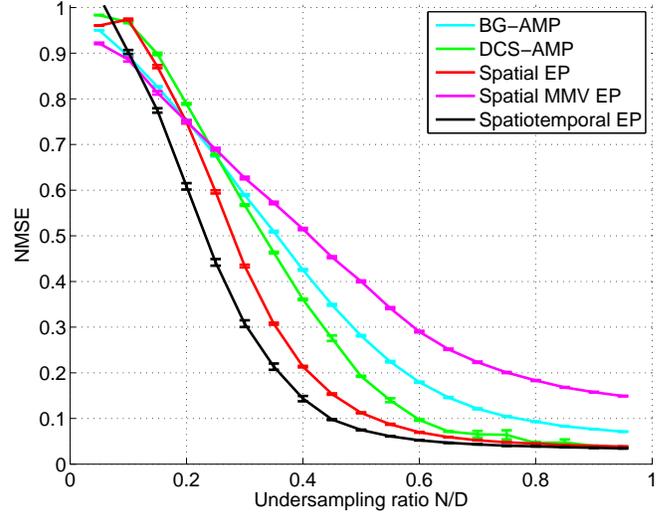}
%}
%\subfigure[]{\includegraphics[width=0.3\textwidth]{figures/F}}
\caption{Normalized mean square error as a function of undersampling ratio. The data are generated from $\Y = \A\X + \E$ with the sparsity pattern shown in figure \ref{fig:true_support}, where $D = 100, T = 100$ and SNR $= 10dB$. The entries in $\A$ are Gaussian i.i.d, i.e. $A_{i,j} \sim \N\left(0, 1/N\right)$. The results are averaged over 100 realizations.}
\label{fig:NMSE}
\end{figure}

\begin{figure}[!t]
\centering
%\subfigure[]{\includegraphics[width=0.3\textwidth]{figures/NMSE}}
%\subfigure[]{
\includegraphics[width=0.48\textwidth]{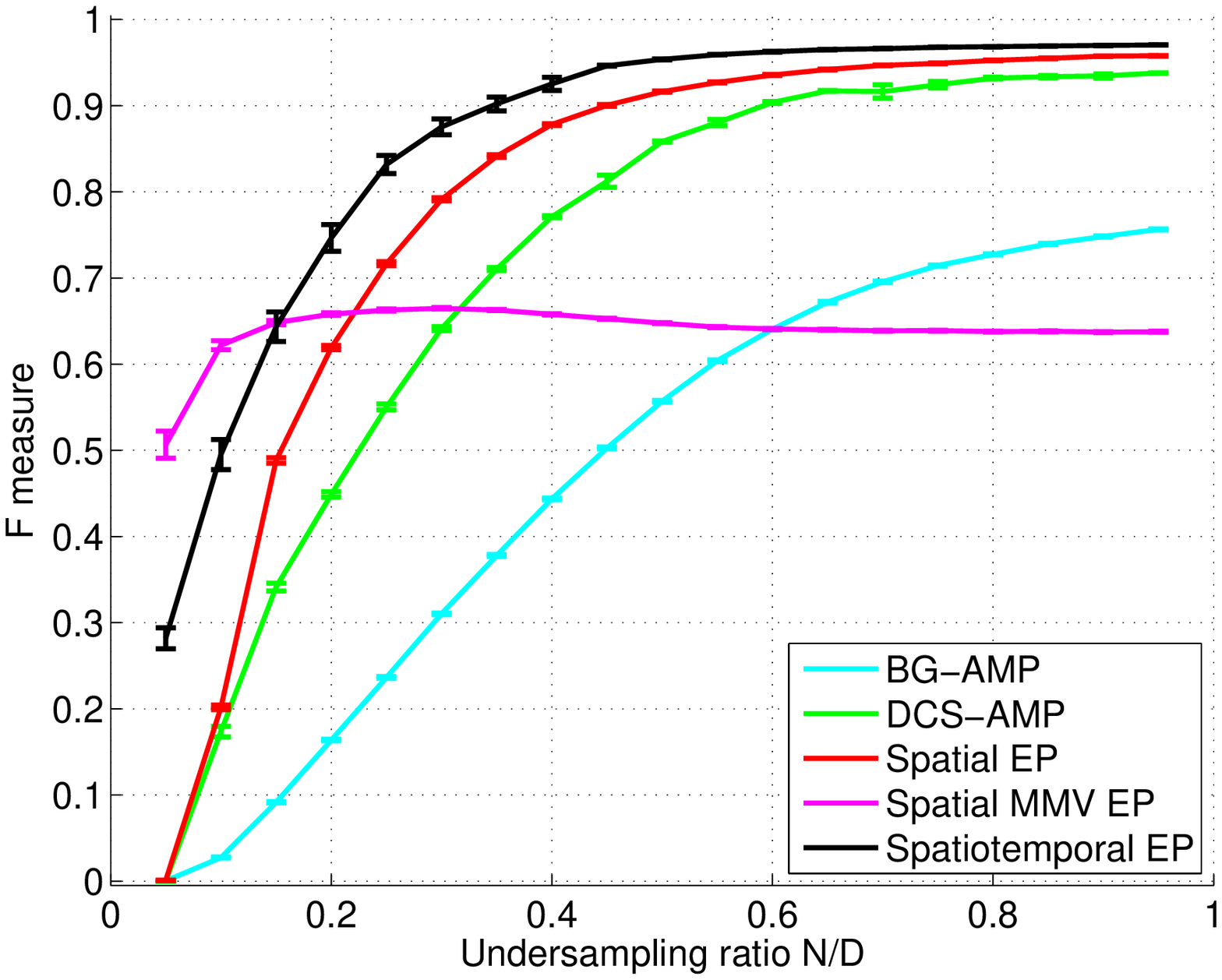}
%}
\caption{F-measure error as a function of undersampling ratio. The data are generated from $\Y = \A\X + \E$ with the sparsity pattern shown in figure \ref{fig:true_support}, where $D = 100, T = 100$ and SNR $= 10dB$. The entries in $\A$ are Gaussian i.i.d, i.e. $A_{i,j} \sim \N\left(0, 1/N\right)$. The results are averaged over 100 realizations.}
\label{fig:F}
\end{figure}

\begin{figure*} 
\centering
\subfigure[]{\includegraphics[width=0.32\textwidth]{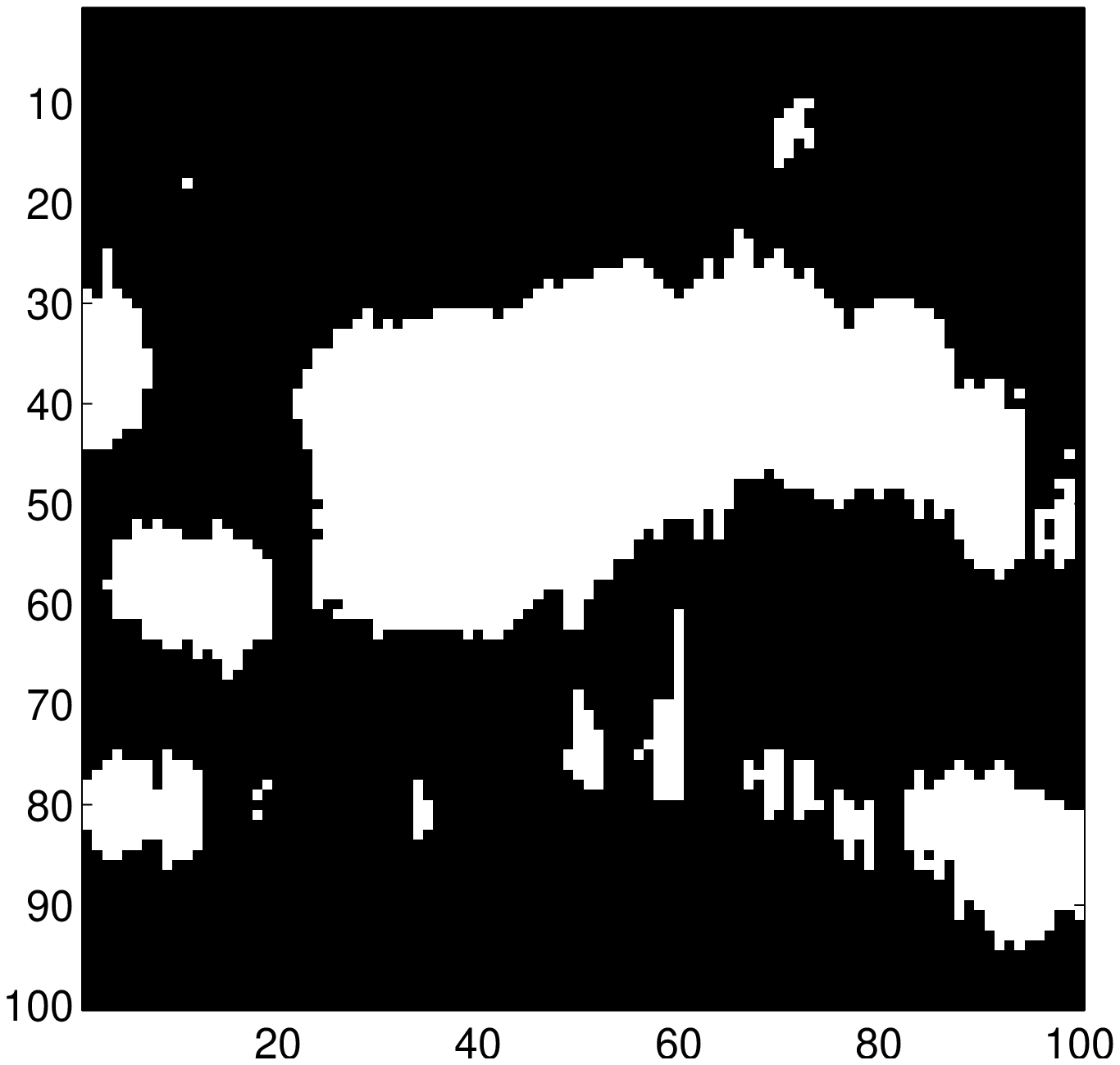}\label{fig:true_support}}
\subfigure[]{\includegraphics[width=0.32\textwidth]{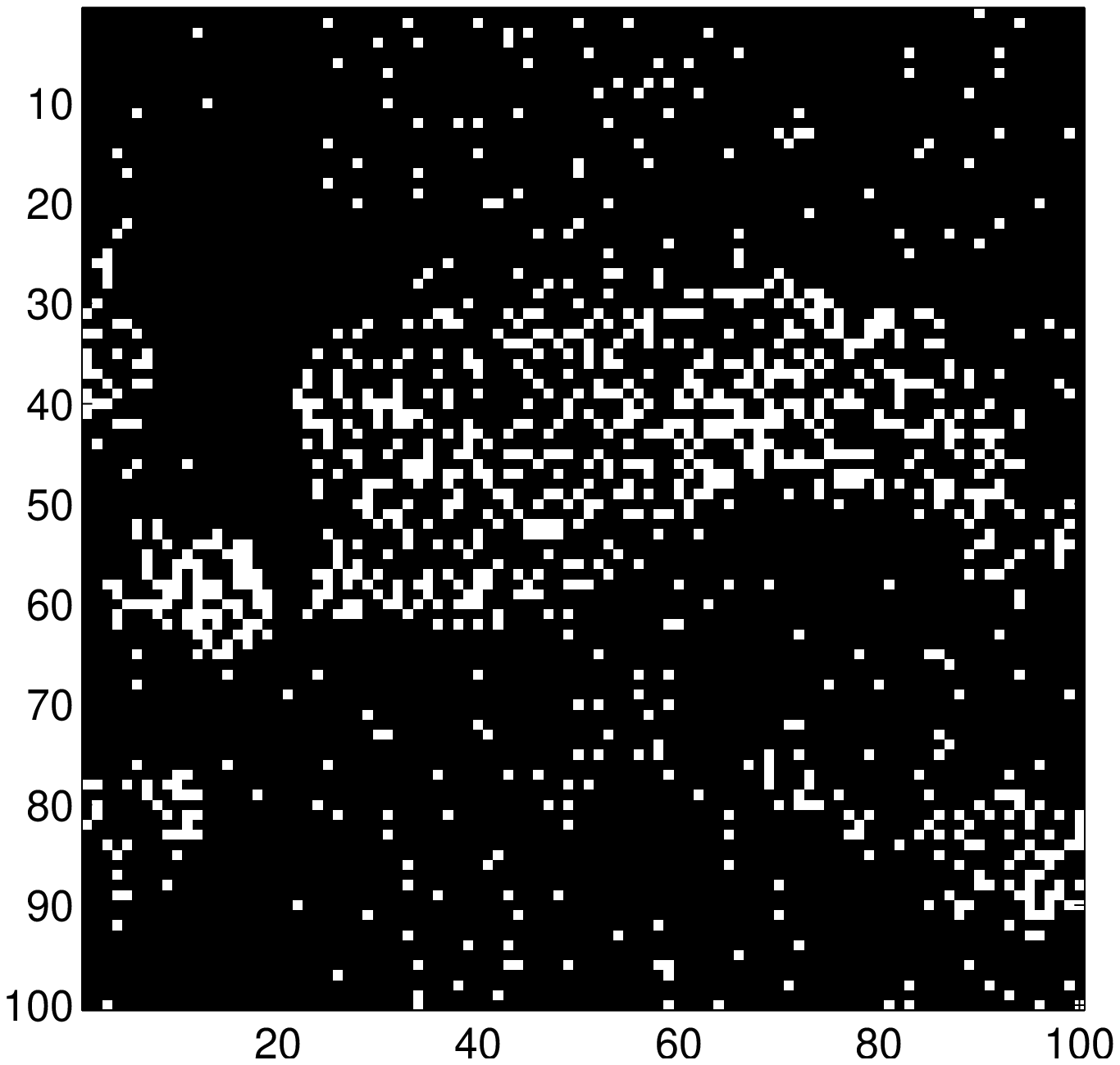}\label{fig:BG-AMP}}
\subfigure[]{\includegraphics[width=0.32\textwidth]{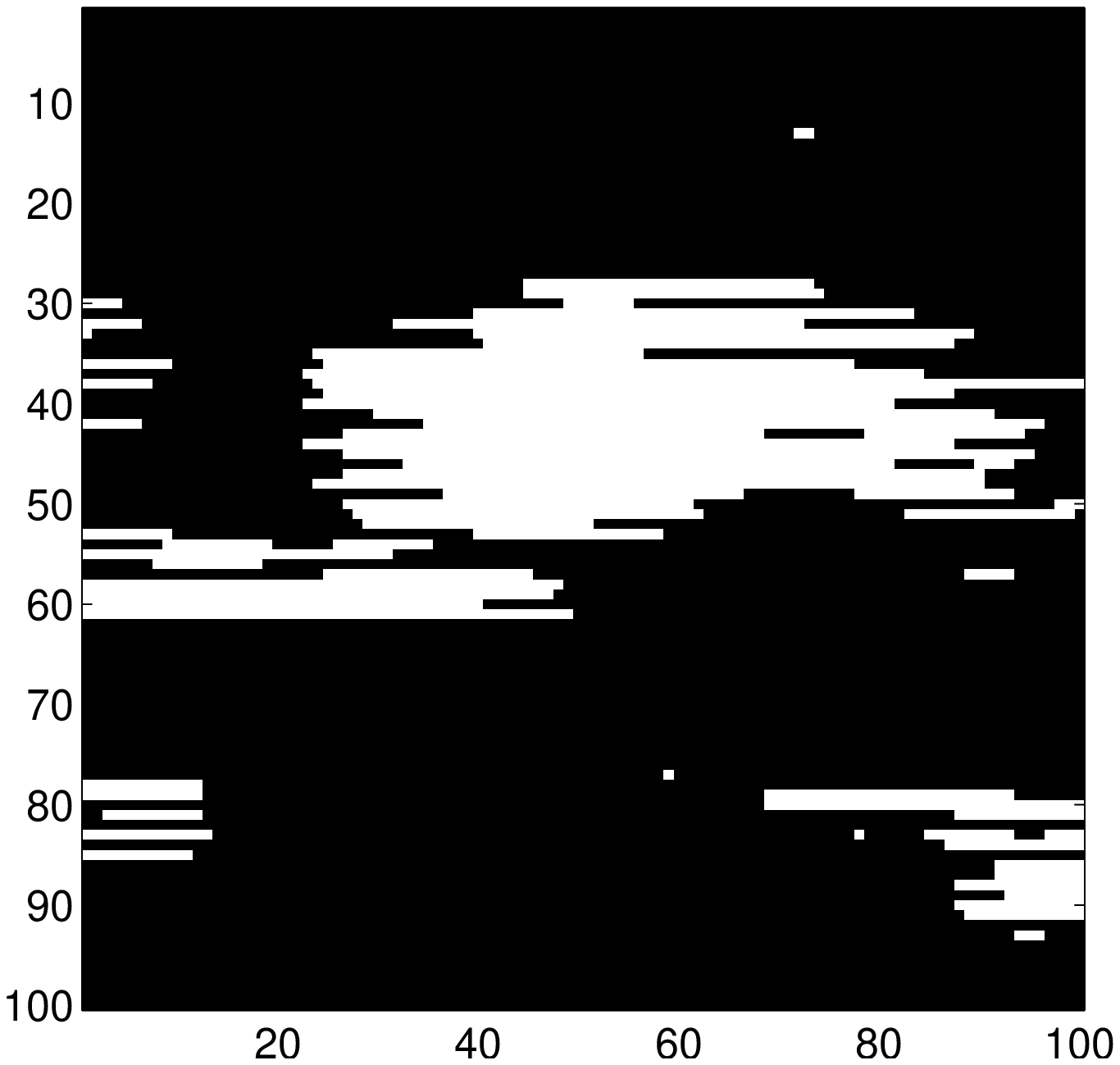}}
\subfigure[]{\includegraphics[width=0.32\textwidth]{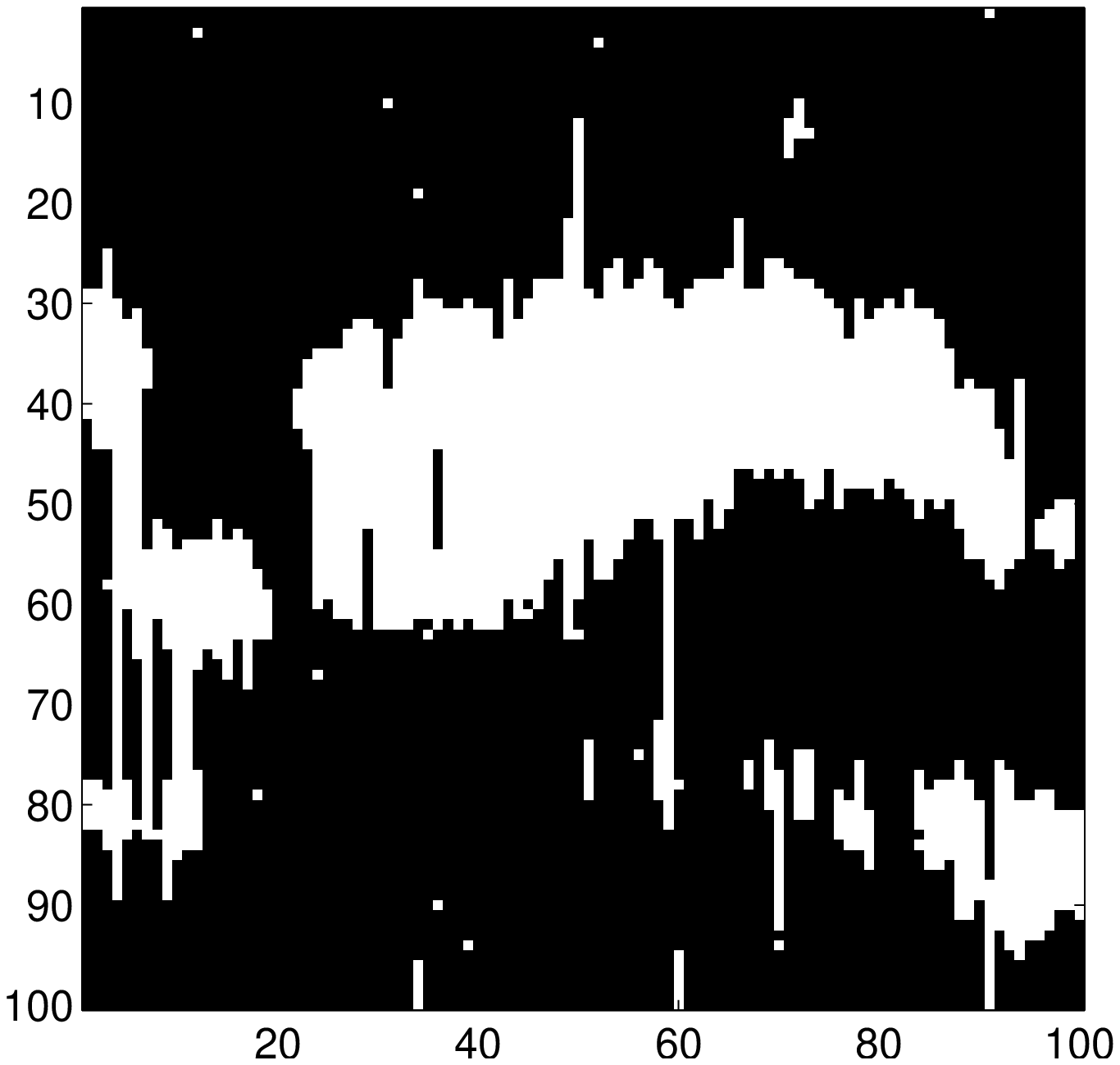}}
\subfigure[]{\includegraphics[width=0.32\textwidth]{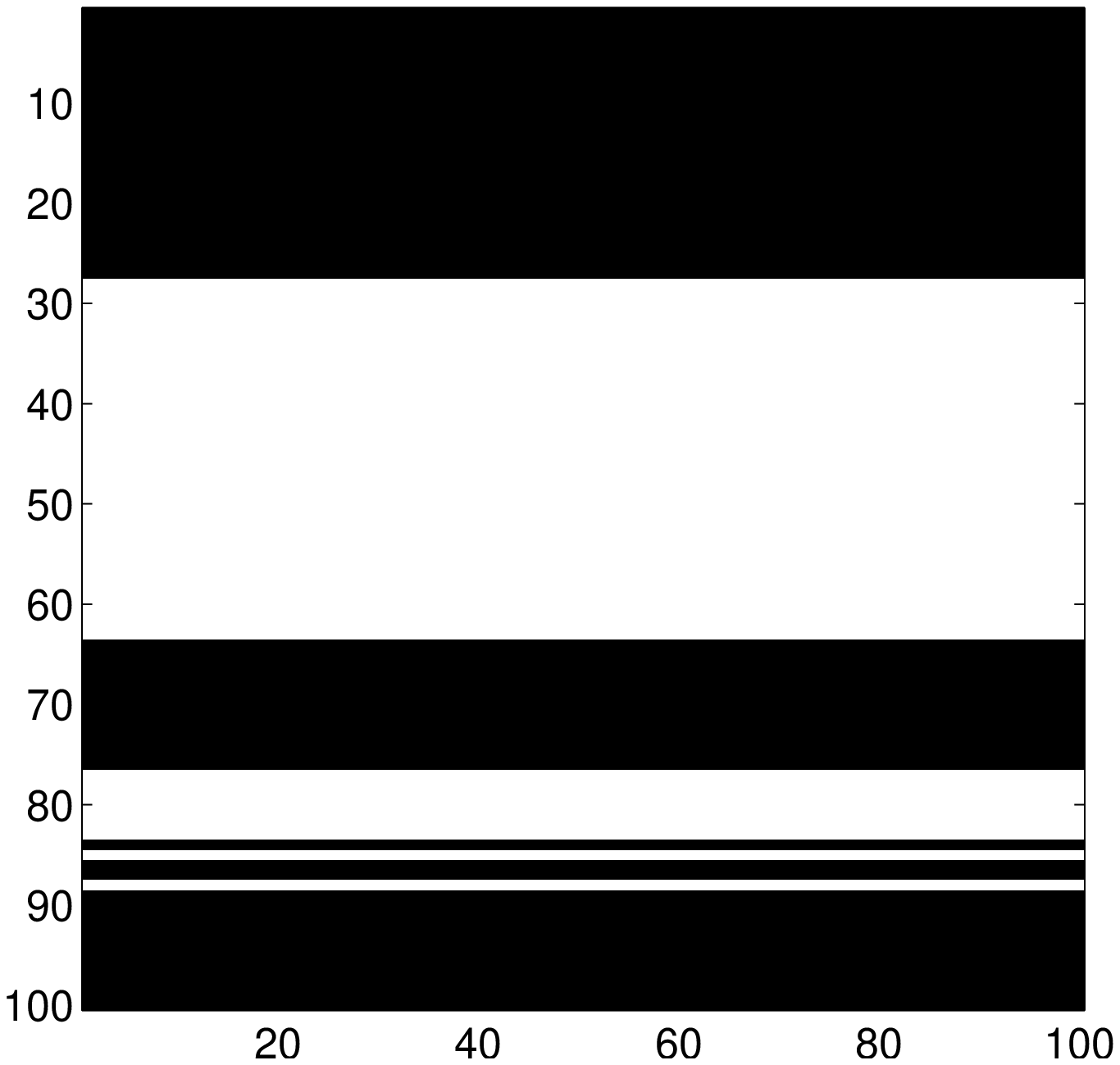} \label{fig:Joint_sparsity}}
\subfigure[]{\includegraphics[width=0.32\textwidth]{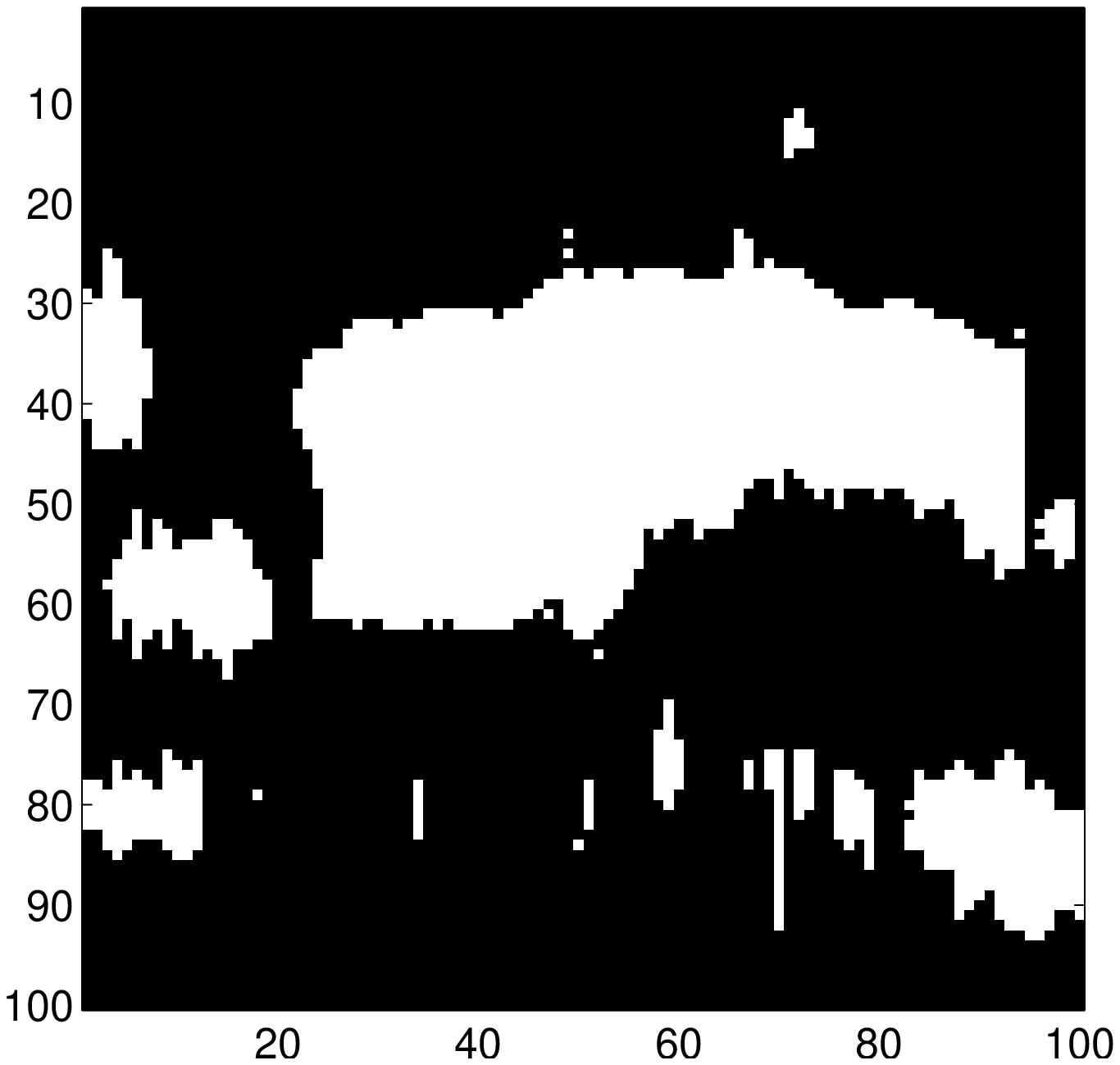}\label{fig:spatiotemporal}}
\caption{True and reconstructed support for the 5 considered methods. The undersampling ratio is $N/D = 0.4$ and $D = 100, T = 100$ and $SNR = 10$dB. a) True support, b) BG-AMP (NMSE = 0.805, F = 0.450), c) DCS-AMP (NMSE = 0.777, F = 0.763), d) Spatial EP (NMSE = 0.658, F = 0.902), e) Spatial MMV EP (NMSE = 0.833, F = 0.663), f) Spatio-temporal EP (NMSE = 0.618, F = 0.935).  }
\end{figure*}

To evaluate the proposed method, we have compared the method to several related solvers: BG-AMP\footnote{We used the implementation in GAMP-toolbox by Sundeep Rangan et al: \url{http://gampmatlab.wikia.com/wiki/}} \cite{vila2013a}, DCS-AMP\footnote{We used the implementation in the DCS-AMP-toolbox by Justin Ziniel: http://www2.ece.ohio-state.edu/~zinielj/dcs/} \cite{ziniel2013a}, Spatial EP (implements the structured spike and slab prior) \cite{NIPS2014_5464} and Spatial MMV EP. The BG-AMP method combines the conventional spike and slab prior with approximate message passing-based  \cite{rangan2011a} inference. We include this method to have a baseline result without any structural assumptions on the sparsity pattern. The DCS-AMP can be seen as an extension of BG-AMP, which assumes that the sparsity pattern evolves slowly in time according to a binary Markov chain. The Spatial EP method assumes spatial correlation in the sparsity pattern, but no temporal correlation. Finally, the Spatial MMV method is similar to Spatial EP but with static sparsity across time, i.e. it assume joint sparsity across time.
\\
\\
To set up the first test we first sampled one realization of $\Z$ using eq. \eqref{eq:prior_z}-\eqref{eq:prior_gamma} with $D = 100$, $T = 100$, $\alpha = 0.99$ and $\beta = 1-\alpha^2$, see figure \ref{fig:true_support}. The average number of non-zero weights per column is fixed to 20. We note that the resulting sample exhibits the spatio-temporal structure as desired. Afterwards, we sample the nonzero coefficients in $\X$ from a standard normal distribution and from these we generate compressive measurements using eq. \eqref{eq:linearinverseproblem}, where $A_{ij} \sim \N\left(0, 1/N\right)$, the SNR $ =10$dB and the undersampling ratio $N/D$ is varied from from $0.05$ to $0.95$. To quantify the performance of the methods we use Normalized Mean Square Error (NMSE) between the true $\X$ and the estimated $\hat{\X}$ given by
\begin{align}
NMSE = \frac{\sum_{i,t} \left(X_{i,t} - \hat{X}_{i,t}\right)^2}{\sum_{i,t} X_{i,t}^2}.
\end{align} 
Furthermore, we evaluate each method's ability to recover the true support $\Z$ using the F-measure \cite{Rijsbergen:1979:IR:539927} based on a MAP estimate of the support $\hat{\Z}$,
\begin{align}
F = 2\frac{\text{precision}\cdot\text{recall}}{\text{precision}+\text{recall}}.
\end{align}
The results are averaged over 100 realizations of the noise $\E$ and non-zero coefficients in $\X$ and are shown in figures \ref{fig:NMSE}-\ref{fig:F}. It is seen that the proposed spatio-temporal method outperforms the other methods both in terms of NMSE and F-measure, but in general it is seen that richer prior assumptions on the support improves the results significantly. We also note that for very undersampled problems, the Spatial MMV EP method with static sparsity actually performs best. But as the undersampling ratio increases, all the other methods, including BG-AMP, outperforms it due to the very high bias of the model.
\\
\\
Figures \ref{fig:BG-AMP}-\ref{fig:spatiotemporal} shows the reconstructed support sets for the undersampling ratio $N/D = 0.4$. It is seen that DCS-AMP and Spatial EP, which models temporal and spatial structure, respectively, clearly outperforms BG-AMP. Furthermore, it is also seen that joint sparsity assumption (fig. \ref{fig:Joint_sparsity}) is too restrictive for these kinds of signals. Again, we note that the spatio-temporal model gives superior results in terms of both F-measure and NMSE. 

\subsection{Experiment 2}
\begin{figure}
\centering
\includegraphics[width=0.48\textwidth]{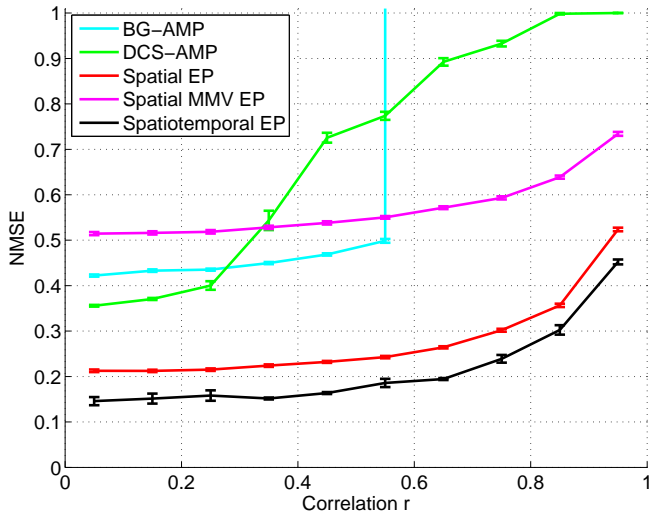}
\caption{NMSE error as a function of undersampling ratio. The data are generated from $\Y = \A\X + \E$ with the sparsity pattern shown in figure \ref{fig:true_support}. The correlation of the $i$'th and $j$'th column of $\A$ is given by $r^{\left|i-j\right|}$. The results are averaged over 50 realizations.}
\label{fig:NMSE2}
\end{figure}

\begin{figure}
\centering
\includegraphics[width=0.48\textwidth]{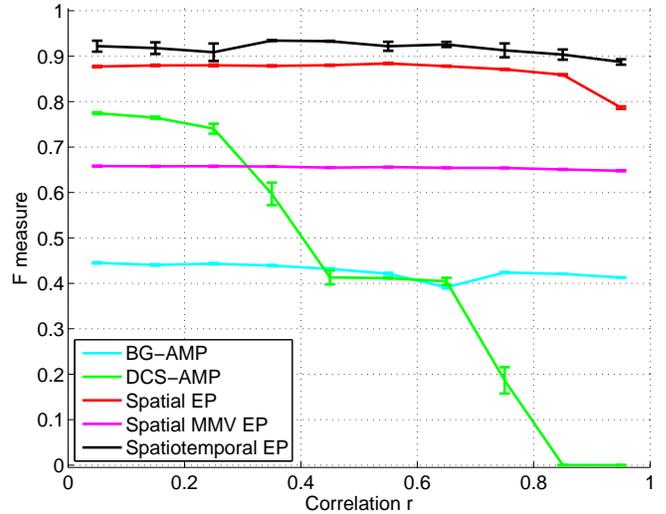}
\caption{F-measure error as a function of undersampling ratio. The data are generated from $\Y = \A\X + \E$ with the sparsity pattern shown in figure \ref{fig:true_support}. The correlation of the $i$'th and $j$'th column of $\A$ is given by $r^{\left|i-j\right|}$. The results are averaged over 50 realizations.}
\label{fig:F2}
\end{figure}

The forward model $\A$ in the EEG source localization problem contains highly correlated columns, i.e. $\A$ is coherent. Therefore, it is of interest to investigate the proposed algorithm's robustness to coherent forward models. The set-up in this experiment is basically the same as for the first experiment, except that undersampling ratio is now fixed to $N/D = 0.4$ and the elements in the forward model $A_{ij}$ are no longer Gaussian i.i.d. Instead we sample the rows of $\A$ from a correlated multivariate normal distribution, such that the columns of $\A$ will be correlated. In particular, the correlation of the $i$'th and $j$'th column of $\A$ is given by $r^{\left|i-j\right|}$. We compute the NMSE and F-measure as a function of the correlation $r$. Note that the BG-AMP and DCS-AMP methods are designed for Gaussian i.i.d forward. These two methods are therefore not expected to perform well in this experiment, but we include them for completeness. The results are averaged over 50 realizations and are shown in figures \ref{fig:NMSE2} and \ref{fig:F2}. The EP-based methods show some robustness to correlation in the columns of $\A$, but the performance does degrade gradually when we increase the correlation. In particular, when changing the correlation $r$ from $0.05$ to $0.95$, the F-measure for the spatio-temporal method only drops from approximate 0.92 to 0.89, but the NMSE increases from approximately 0.15 to 0.45.

\section{Conclusion \& outlook}
We extended the structured spike and slab prior and the associated Expectation Propagation inference scheme to cope with smooth temporal evolution of the sparsity pattern. Based on numerical experiments with synthetic data we demonstrated the benefits of the extended model. In particular, we showed that the method outperformed the reference methods. Future work includes developing an automated approach learning the hyperparameters of the prior and applying the proposed method to a real EEG source localization problem.

\section*{Acknowledgment}
The authors would like to thank Sundeep Rangan et al. and Justin Ziniel for making their toolboxes available online.
% trigger a \newpage just before the given reference
% number - used to balance the columns on the last page
% adjust value as needed - may need to be readjusted if
% the document is modified later
%\IEEEtriggeratref{8}
% The "triggered" command can be changed if desired:
%\IEEEtriggercmd{\enlargethispage{-5in}}

% references section

% can use a bibliography generated by BibTeX as a .bbl file
% BibTeX documentation can be easily obtained at:
% http://www.ctan.org/tex-archive/biblio/bibtex/contrib/doc/
% The IEEEtran BibTeX style support page is at:
% http://www.michaelshell.org/tex/ieeetran/bibtex/
%\bibliographystyle{IEEEtran}
% argument is your BibTeX string definitions and bibliography database(s)
%\bibliography{IEEEabrv,../bib/paper}
%
% <OR> manually copy in the resultant .bbl file
% set second argument of \begin to the number of references
% (used to reserve space for the reference number labels box)
\bibliographystyle{IEEEtran}
\bibliography{spars}

% that's all folks
\end{document}